\documentclass[journal]{IEEEtran}
\usepackage{color,xcolor}
\usepackage{graphicx}
\usepackage{subfigure}
\newcommand{\system}{FT-TDR\xspace}
\newcommand*{\eg}{\emph{e.g.}\@\xspace}

\usepackage{multirow}
\usepackage{amsthm,amsmath,amssymb}
\usepackage{mathrsfs}
\usepackage{diagbox}
\usepackage[ruled]{algorithm2e}
\usepackage{bm}
\usepackage{booktabs}
\usepackage{soul}
\soulregister\cite7 
\soulregister\citep7 
\soulregister\citet7
\soulregister\ref7 
\soulregister\pageref7
\soulregister\system7

\usepackage[inline]{enumitem}

\ifCLASSINFOpdf
\else
\fi
\begin{document}
%
\title{FT-TDR: Frequency-guided Transformer and Top-Down Refinement Network \\ for Blind Face Inpainting}
%
%
%

\author{Junke~Wang, 
        Shaoxiang~Chen, 
        Zuxuan~Wu, 
        and~Yu-Gang~Jiang
\thanks{Junke Wang, Shaoxiang Chen, Zuxuan Wu and Yu-Gang Jiang are with Shanghai Key Lab of Intelligent Information Processing, School of Computer Science, Fudan University and Shanghai Collaborative Innovation Center on Intelligent Visual Computing. \\ 
email: wangjk21@m.fudan.edu.cn, \{sxchen13, zxwu, ygj\}@fudan.edu.cn\\
Corresponding author: Yu-Gang Jiang. 
}

}

\maketitle

\begin{abstract}
Blind face inpainting refers to the task of reconstructing visual contents without explicitly indicating the corrupted regions in a face image. Inherently, this task faces two challenges: (1) how to detect various mask patterns of different shapes and contents; (2) how to restore visually plausible and pleasing contents in the masked regions. In this paper, we propose a novel two-stage blind face inpainting method named Frequency-guided Transformer and Top-Down Refinement Network (\textbf{\system}) to tackle these challenges. Specifically, we first use a transformer-based network to detect the corrupted regions to be inpainted as masks by modeling the relation among different patches. For improved detection results, we also exploit the frequency modality as complementary information and capture the local contextual incoherence to enhance boundary consistency. Then a top-down refinement network is proposed to hierarchically restore features at different levels and generate contents that are semantically consistent with the unmasked face regions.
  Extensive experiments demonstrate that our method outperforms current state-of-the-art blind and non-blind face inpainting methods qualitatively and quantitatively.
\end{abstract}


%
\IEEEpeerreviewmaketitle

\section{Introduction}
%
%
%
%
\IEEEPARstart{F}{ace} image inpainting aims to reconstruct the missing parts of the input face image based on valid contexts. It can be applied to various multimedia tasks, such as image restoration and face attribute editing.  Generally, most existing face inpainting methods \cite{li2017generative,liu2018image,nazeri2019edgeconnect,yu2019free,yang2019lafin} require to take both the corrupted image and its corresponding mask as input. However, in most realistic scenarios, it is impractical to obtain the masks directly, and manual labeling is often time-consuming and inaccurate. Recently, \cite{wang2020vcnet} considered a new task to restore contents without specifying masks that indicate missing areas in an image, named \textbf{blind image inpainting}.

\begin{figure}[t]
\centering
  \includegraphics[width=\linewidth]{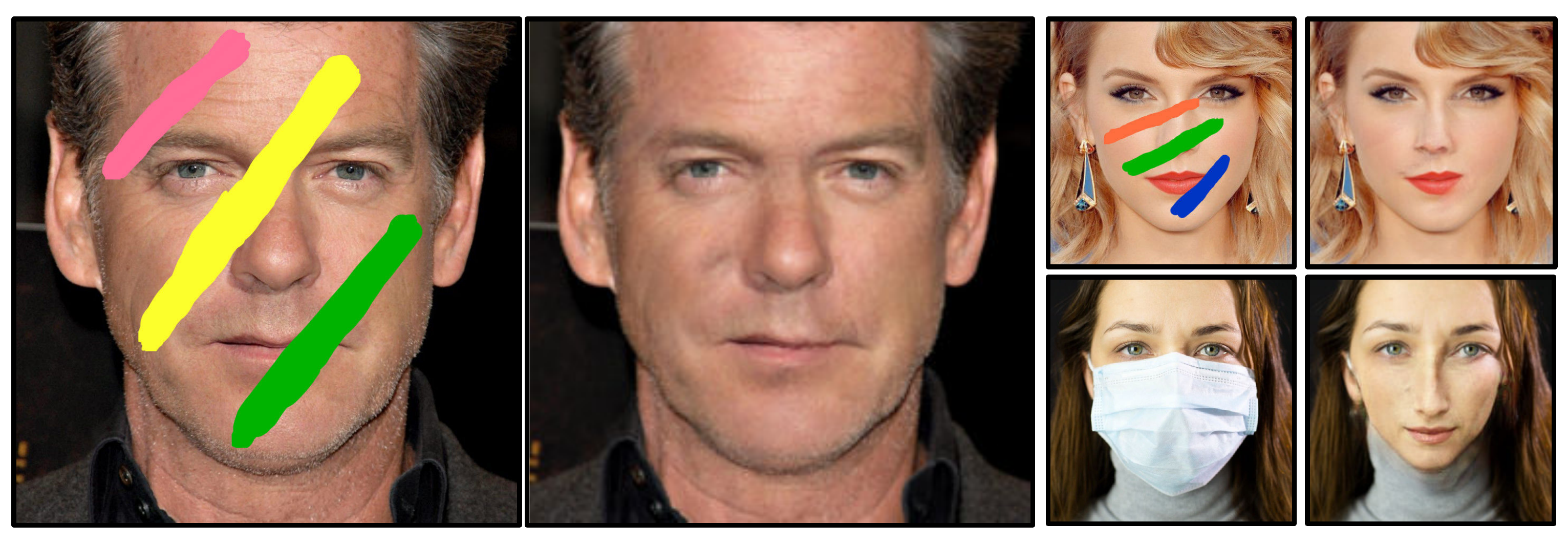}
  \caption{Face inpainting results by our proposed blind face inpainting method on faces images occluded by graffiti and masks. These images are collected from the Internet, the filling patterns of which are not included in our training process. In addition, no masks are provided during inference.}
  \label{fig:teaser}
\end{figure}

In this paper, we adopt the blind inpainting settings from \cite{wang2020vcnet} and focus on a more specific task: blind face inpainting. Compared with natural scenes, human faces have more complex structures and finer textures, thus requiring higher visual quality for the restoration results and is thought to be more challenging \cite{li2017generative, yang2019lafin}.
Specifically, the input to our method is a corrupted face image that could be contaminated by various patterns. We aim to recognize the visually unreasonable regions in the input image and complete natural and pleasing contents within these regions. Figure~\ref{fig:teaser} shows several inpainting results by our method on real cases, e.g., faces images occluded by graffiti and masks. 

The challenges of blind face image inpainting are twofold. The first is how to detect various damage patterns. \cite{wang2020vcnet} uses a naive convolutional encoder-decoder architecture, where the encoder is for semantic feature extraction and the decoder is for pixel-wise classification. However, due to the limitation of convolution operations, such a structure may struggle to handle the long-range relation, thus failing to detect complex corruption patterns. We argue that modeling spatial long-range information is necessary to globally integrate the features of different regions and is crucial to recognizing corrupted regions in this task. 
In addition to long distance relation modeling for global information integration, local contextual information is of significant importance to enhance the boundary consistency of the prediction results. Based on the fact that visually abnormal areas are usually inconsistent with the surrounding contexts, pair-wise similarity of local patches can be utilized to effectively capture the inconsistency at the boundary. 
The patterns of corrupted regions in real scenes could be diverse, making global information modeling and local feature extraction in the RGB domain not adequate to detect some subtle damaged regions. Prior studies \cite{yu2019attributing, durall2019unmasking} in Deepfake detection suggest that the artifacts of forged images can be perceived in the frequency domain in the form of unusual frequency distributions. Inspired by this, we further exploit the frequency modality for our mask detection.

Based on the above motivations, we first propose a Transformer-based Mask Detection Module for corrupted region prediction, using the self-attention mechanism to model the relation of different local regions. 
It is widely recognized that transformer-based architecture has recently demonstrated superior performance on a broad range of vision and language tasks \cite{dosovitskiy2020image, carion2020end, fan2021multiscale, wang2021efficient}, mostly because of its strong capabilities in modeling long-range relation. 
To exploit the frequency domain, we also transform the corrupted image into frequency-aware components which preserve abnormal frequency signals based on Discrete Cosine Transform (DCT) and use stacked convolution layers to extract frequency modality features. These features are incorporated into the transformer encoder for information integration.
Furthermore, we propose a Patch Similarity (PS) Block and embed it into the transformer encoder, which explicitly calculates the pair-wise similarity between neighboring local patches to capture the local semantic inconsistency. In general, we capture both \textit{frequency modality anomaly} and \textit{contextual semantic incoherence} based on a global relation-modeling transformer network to detect the corrupted regions on faces. 

The second obstacle of this task is how to restore visual contents that are both consistent with the surrounding context and visually pleasing. To complete both the geometric structure and fine texture of the masked regions, a large group of works~\cite{song2018contextual,ren2019structureflow,yu2019free} use two encoder-decoders to separately learn structural and textural features.
However, the independent learning of structure and texture reconstructions will produce artifacts in the final output. 
To avoid this,~\cite{liu2020rethinking} uses the features from deep layers of the encoder to reconstruct structural semantics and the features extracted from shallow layers to reconstruct textural details. But additional information (i.e., ground-truth structure image generated by an edge-preserving smoothing method RTV \cite{xu2012structure}) is needed to supervise the effective feature extractions of two branches, which will greatly increase the complexity of the model.

To address the above issues, we propose a Top-Down Refinement (TDRefine) Module which consists of a bottom-up path and a top-down path. The bottom-up path captures rich textural information from low-level features and structure knowledge from high-level features. Then in the top-down path, the encoded structural features are merged with the low-level features by the top-down refinement fusion lock. In this way, the texture and structure information is jointly utilized in a single network. 

In summary, our proposed method addresses the technical challenges of blind face image inpainting following a two-stage pipeline. First, it can detect the corrupted areas with decent performance even when the corruption patterns are unseen to the trained model. Second, it can generate visually reasonable and pleasing contents within the predicted or given masked regions.

Our contributions can be summarized as follows:
\begin{itemize}
\item We propose a novel Transformer-based Mask Detection Module to detect the corrupted regions based on both frequency modality anomaly and contextual incoherence, which better utilizes the information contained in face images with a transformer architecture.

\item We design a Top-Down Refinement (TDRefine) Module to restore the hierarchical features of the masked regions implicitly using a top-down refinement architecture, and finally generate both realistic and high-quality images.

\item Extensive experiment results demonstrate our model outperforms previous state-of-the-art non-blind facial inpainting methods both qualitatively and quantitatively.
\end{itemize}

\begin{figure*}[t]
\centering
\includegraphics[width=\textwidth]{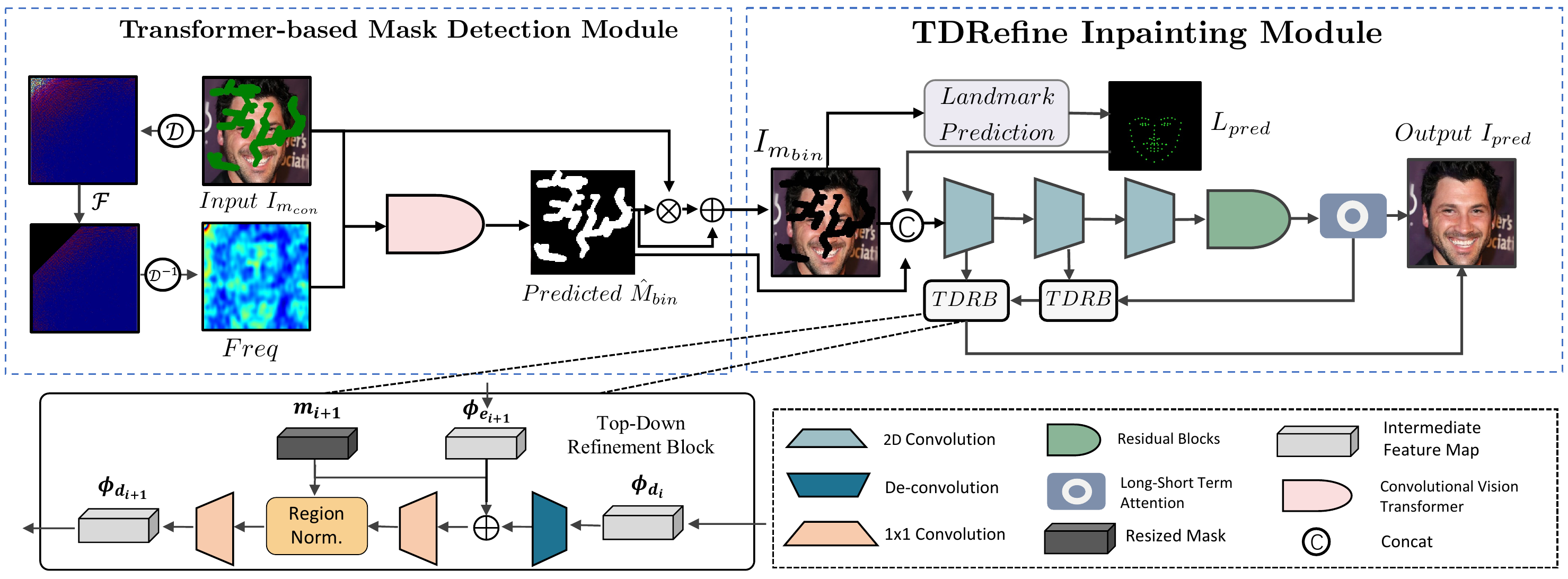}
\caption{Best viewed in color. Overview of the proposed pipeline. For a corrupted face image, a binary mask is first predicted by the mask detection module, then a binary-masked image is generated by combining the corrupted image and predicted binary mask. After that, the landmarks are estimated by the landmark prediction module using the binary-masked image as input. Finally, the inpainting module applies the landmarks as a prior to the binary-masked image to inpaint the face image.}
\label{figure:framework}
\end{figure*}



\section{Related Work}

\subsection{Deep Image Inpainting}
Recently, deep learning based methods have become prevalent in image inpainting. \cite{iizuka2017globally} puts forward an approach which can generate inpainted images that are both locally and globally consistent with the surrounding areas by using a global and local context discriminator. \cite{liu2018structure} formulates the image inpainting problem as an energy optimization problem, and exploits an EM-like approach based on homography transformations to solve it .
\cite{liu2018image,yu2019free} propose novel convolution methods and mask updating mechanisms to make networks adaptive to the masked input. \cite{lahiri2020prior} uses predicted prior to guide the inpainting network for better retention of the structure of the object to be restored. Besides, \cite{zhao2020uctgan, cai2020piigan} explore to produce multiple plausible results for a given masked input based on conditional probability models.

Face inpainting is more challenging compared with general image inpainting tasks because facial attributes have strong visual consistency to preserve and contain large appearance variations. \cite{song2019geometry} uses estimated facial landmark heatmaps and parsing maps to guide a generator of encoder-decoder structure to complete a face image conditioned on both the uncorrupted regions and the estimated facial geometry images. However, \cite{yang2019lafin} argues that redundant face geometry like parsing maps may degenerate the performance when feeding slightly inaccurate information into the inpainting module, instead, they use facial landmarks as the indicator to reconstruct the missing regions. Although these prior guided methods could recover natural contents, the synthesized faces still lack of high-frequency details. To address this problem, \cite{wang2020recurrent} proposes a recurrent generative adversarial network to hierarchically restore the textures within masked regions. \cite{wang2019laplacian} utilizes a Laplacian pyramid adversarial network to complete the multi-scale information of missing face regions in a coarse-to-fine manner.

\subsection{Blind Image Inpainting} 
Existing blind image inpainting works \cite{cai2017blind,liu2019deep} are based on the assumption that the corrupted areas are filled with simple data distributions, such as thin stroke masks filled with constant values or regular masks filled with Gaussian noise. This setting is different from most scenarios in real life, which limits the application scope of the proposed approach. In addition, the pixels in the masked area are significantly different from those in other areas, which makes the network easily develop the capability to identify abnormal areas and overfit to the specific mask patterns. Comparatively, we adopt more complex mask filling patterns that are closer to the real-life data distributions. Recently, \cite{wang2020vcnet} proposes a novel data generation strategy to enrich the training data as much as possible, and formulate the versatile blind inpainting task. Following \cite{wang2020vcnet}, we additionally incorporate the frequency modality and contextual incoherence into mask prediction, and focus on the more challenging face inpainiting task.

\subsection{Visual Transformer}
Exemplary performance of Transformer models \cite{vaswani2017attention} in natural language processing have intrigued the vision community to apply them in vision problems. \cite{chen2020generative, dosovitskiy2020image} purely use transformer for image classification. \cite{carion2020end, zhu2020deformable} use the self-attention mechanism in the transformer to enhance the specific modules of traditional object detectors. \cite{wang2021m2tr} proposes to use a multi-scale transformer
to detect local inconsistency in forged images at different scales.  The above methods extract the features of input images through transformer encoders, and output low-dimensional predictions. Comparatively, \cite{zeng2020learning} proposes a spatial-temporal transformer network for video inpainting, while \cite{zheng2020rethinking} adopts a pure transformer to encode the image as a sequence of patches and further predict the segmentation map by a decoder. In this paper, we propose a novel Transformer-based Mask Detection Module to detect the damaged regions of face images, guided by the frequency modality features.

\section{METHODOLOGY}
In this section, we introduce our proposed blind face inpainting method named \system. 
As shown in Figure~\ref{figure:framework}, it consists of two parts, i.e., Transformer-based Mask Detection Module and TDRefine Inpainting Module.

\subsection{Training Data Generation}
Let $I_{gt}$ be the uncorrupted ground truth face image. Under blind inpainting setting, we generate the masked image for training $I_{m_{con}}$ following the strategy proposed by \cite{wang2020vcnet}:
\begin{equation}
    I_{m_{con}} = I_{gt} \odot (1-M_{bin}) + M_{con} \odot M_{bin},
\end{equation}
where $M_{bin}$ is a binary mask (with value 0 for valid pixels and 1 for otherwise), $M_{con}$ is a noisy visual signal (i.e., constant value or real-world images), and $\odot$ is Hadmard product operator. Note that during the process of training, both $M_{bin}$ and $M_{con}$ are randomly selected for $I_{gt}$ and there is no correspondence between them.

\subsection{Transformer-based Mask Detection Module}
The target of our Transformer-based Mask Detection Module (TMDM) $\mathcal{G}_{M}$ (parameterized by $\theta_{m}$) is to recognize the visually abnormal areas on a face image and predict a binary mask $\hat{M}_{bin}$:
\begin{equation}
\hat{M}_{bin}=\mathcal{G}_{M}(I_{m_{con}}; \theta_{m}).
\end{equation}

Then a binary-masked face image $I_{m_{bin}}$ is obtained by blending the corrupted image $I_{m_{con}}$ and the predicted mask $\hat{M}_{bin}$: 
\begin{equation}
    I_{m_{bin}} = I_{m_{con}} \odot (1-\hat{M}_{bin}) + \hat{M}_{bin}.
\end{equation}

\begin{figure}[t]
  \centering
  \includegraphics[width=1.0\linewidth]{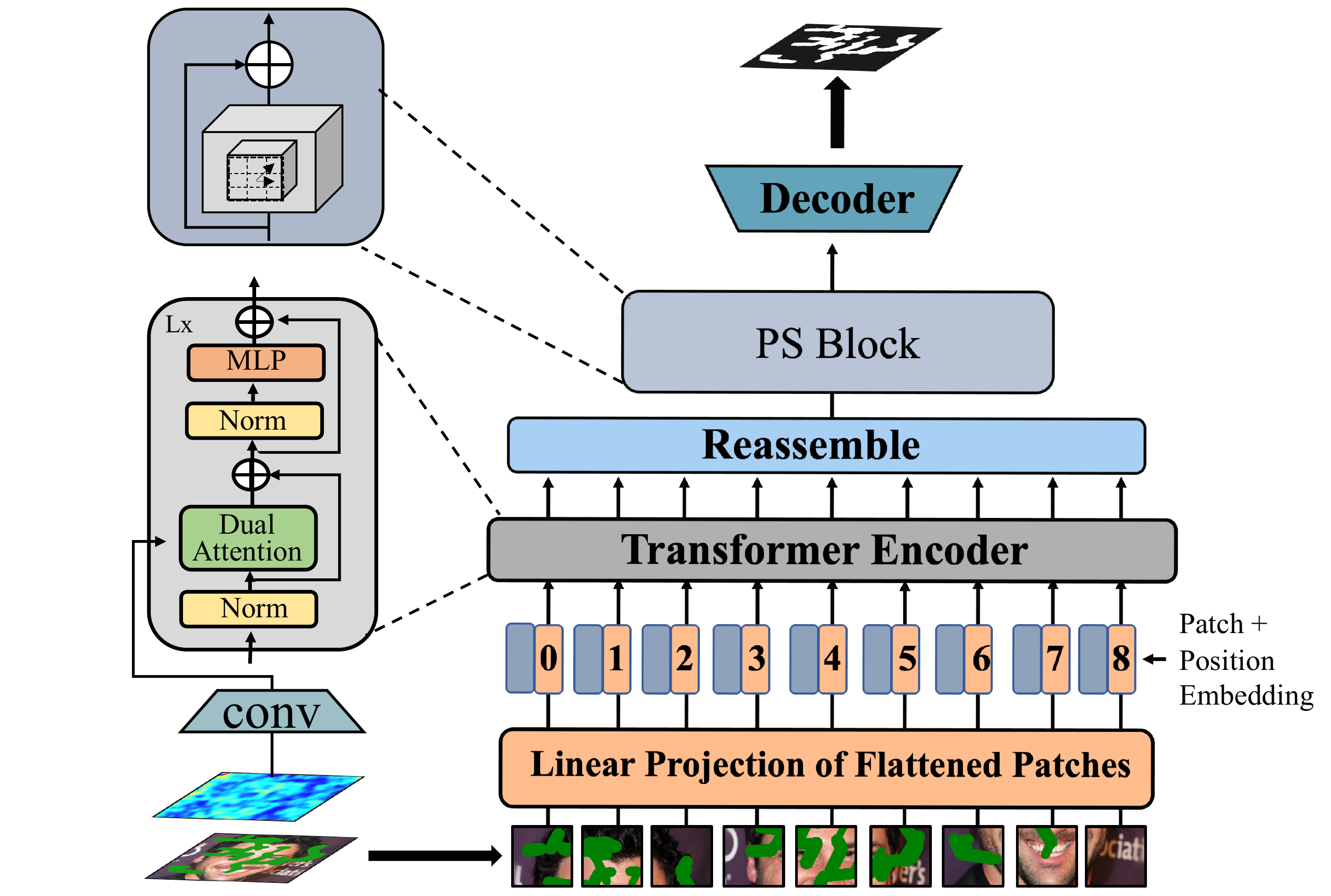}
  \caption{Illustration of the proposed Transformer-based Mask Detection Module. The corrupted image is split into a sequence of patches and input to a convolutional vision transformer for feature extraction. The high-pass frequency-aware representations are passed through several convolution layers to obtain the frequency attention, which are then concatenated with the self-attention in the transformer to form a dual-path attention.}
  \label{figure:tmdm}
\end{figure}

Specifically, TMDM consists of two components: a Frequency Anomaly Detector (FAD) and a Convolutional Vision Transformer (CViT). The detailed architecture is illustrated in Figure~\ref{figure:tmdm}.

\subsubsection{Frequency Anomaly Detector}
Considering DCT has the property that the high and low frequency components of the resulting signals distribute in different locations, we first apply DCT to transform $I_{m_{con}}$ $\in$ $\mathbb{R}^{H \times W \times 3}$ from RGB domain to frequency domain and use a hand-crafted filter \cite{chen2021local} to filter out the low-frequency information and amplify visually unreasonable signals:
\begin{equation}
    D_{high} = \mathcal{F}(DCT(I_{m_{con}}), \alpha),
\end{equation}
where $\mathcal{F}$ is a high-pass filter, $\alpha$ is the manually-chosen threshold which controls the frequency components to be filtered out, and $D_{high}$ $\in$ $\mathbb{R}^{H \times W \times 1}$ is the high frequency component of the input corrupted image $I_{m_{con}}$.
Then to preserve shift invariance and local consistency of natural images and use the representation learning capability of CNN to extract features, we then invert the filtered signals back into RGB space via inverse DCT to obtain the frequency-aware representation: $F = DCT^{-1} (D_{high})$, where $F$ $\in$ $\mathbb{R}^{H \times W \times 1}$.

\subsubsection{Convolutional Vision Transformer} As previously discussed, we process the mask detection input in a sequence-to-sequence manner to capture the relation of different regions.
First, we reshape the input image $I_{m_{con}}$ into a sequence of flattened patches and embed them into 1D feature embeddings $\{z_{i} \}_{i=1}^{P}$, where $z_{i} \in \mathbb{R}^{1 \times Q}$ and $P$ is the length of sequence. 
In this paper, we set $P$ to $64$, and $Q$ to $(H/8) \times (W/8) \times 64$.
Then we add position embeddings to these features to obtain feature vectors for each patch, and input them to $L$ stacked transformer encoders. Each encoder layer has a standard architecture that consists of a multi-head attention block and a multi-layer perceptron.
Specifically, For each head $h$, we use fully connected layers to map the feature vectors into query, key, and value embeddings $q^{h}$, $k^{h}$, and $v^{h}$ respectively. Then matrix multiplication and $softmax$ function are implemented to calculate the attention $A^{h}$ for different heads :
\begin{equation}
A_{i,j}^{h} = Softmax(\frac{q_{i}^h k_{j}^{h\top}}{\sqrt{d_{k}}}),
\end{equation}
where i,j is the position index. Then the resulting attention maps of different heads are concatenated along the first dimension to obtain the final self-attention map $A$ $\in$ $\mathbb{R}^{N_h \times P \times P}$, and $N_h$ is the number of heads. Then we use several convolutional layers to encode the frequency-aware representation $F$ into a frequency modality attention map $A_{frep}$  $\in$ $\mathbb{R}^{C \times P \times P}$. It is then fused with the attention map of the visual features so that complementary information of the frequency modality can be utilized to better recognize corrupted regions.
\begin{equation}
     A_{dual} = Softmax(conv_{1 \times 1} ([A, A_{freq}])).
\end{equation}
where $[,]$ denotes concatenating along the first dimension and $conv_{1 \times 1}$ denotes the $1 \times 1$ convolution that squeezes the number of channels back to $N_h$.
With the dual attention map $ A_{dual}$, we obtain the output for each query by computing the weighted summation of the attention weights and values $v^{h}$ of relevant patches. 
The outputs are added to the input feature embeddings by a residual connection, and then fed to a MLP.
With $L$ stacked transformer encoders, we obtain the features that are aware of the region-wise relation and sensitive to subtle signals in the frequency domain, we then reassemble it to a 2D feature map $T \in$ $\mathbb{R}^{H/8 \times W/8 \times 64}$.  

In addition, we introduce a Patch Similarity Block (PS Block in Figure~\ref{figure:tmdm}) on top of the transformer.
Based on the fact the visually abnormal regions are usually incoherent with the surrounding context along the edges,
we calculate the similarity between different feature vectors within a local patch to further enhance the edge consistency of the predicted results, and obtain the edge map $E$:
\begin{equation}
\begin{split}
    E_{i} =  \frac{1}{\left| \Omega \right|}\sum_{j \in \Omega} Sim(T_{i}, T_{j}), \\
\end{split}
\end{equation}
where $\Omega$ denotes a small neighboring patch in the feature map $T$ around $i$ (in this paper we set the size of $\Omega$ to 9).
The similarity measurement function $Sim(\cdot)$ that we use is cosine similarity. Then we add the edge map $E$ and the input feature map $T$ to obtain an edge-preserving feature map $F_{edge}$. 

Finally, we use consecutive bilinear upsampling layers and $1 \times 1$ convolutional layers to progressively increase the spatial resolution of $F_{edge}$ and obtain the mask detection results $\hat{M}_{bin}$. Cross-entropy loss and dice loss are combined to supervise the learning of $\mathcal{G}_{M}$.

\subsection{TDRefine Inpainting Module}
Our Top-Down Refinement Inpainting Module $\mathcal{G}_{P}$ follows the encoder-decoder~\cite{johnson2016perceptual} based architecture.
Generally, $\mathcal{G}_{P}$ takes the binary-masked image $I_{m_{bin}}$, the predicted mask $\hat{M}_{bin}$ from the mask detection module as inputs, and outputs a restored image $I_{pred}$:  
\begin{equation}
    I_{pred} = \mathcal{G}_{P}(I_{m_{bin}}, \hat{M}_{bin}; \theta_{p}),
\end{equation}
where $\theta_{p}$ denotes the network parameters. 

Specifically, taking $I_{m_{bin}}$ as inputs, we first adopt the Landmark Prediction Module proposed by~\cite{yang2019lafin} to obtain the predicted facial landmarks $L_{ldmk}$. We further concatenate $L_{ldmk}$ with $I_{m_{bin}}$ and $\hat{M}_{bin}$, and input them to the first TDRB block. \footnote{The effectiveness of landmark detection on corrupted face images is shown in the Supplementary Materials.}. The bottom-up path of the TDRefine module contains an encoder that gradually down-sample twice, followed by 7 residual blocks with dilated convolutions and a long-short term attention block~\cite{zheng2019pluralistic}.
The stacked dilated blocks can enlarge the receptive field, and the long-short attention layer is used to merge the features before and after residual blocks. 

In the top-down path are several Top-Down Refinement Fusion Blocks (TDRB), in addition to up-sampling the feature maps,
the TDRB is also responsible for connecting the decoder layers with the successive encoder layers at different levels, 
so that the low-level texture information can be integrated into the high-level structure information in the top-down path. The TDRB can be formulated as:
\begin{equation}
 \phi_{d_{i+1}} = \texttt{TDRB}(\phi_{d_{i}}, \phi_{e_{i+1}}, m_{i+1}).
\end{equation}
where $\phi_{d_{i}}$ denotes the feature maps generated in the top-down pass, $\phi_{e_{i+1}}$ denotes the feature maps of the encoder layer, and $m_{i+1}$ is the predicted mask indicating the regions to be inpainted. Concretely, we first use de-convolution layers to up-sample $\phi_{d_{i}}$ to the same size as $\phi_{e_{i+1}}$,
and adaptively fuse them according to the mask $m_{i+1}$:
\begin{equation}
 \tilde{\phi}_{d_{i+1}} =  conv_{1\times1}( (deconv(\phi_{d_{i}}) \odot m_{i+1} + \phi_{e_{i+1}} \odot (1-m_{i+1}) ).
\end{equation}
Then we equalize the features inside and outside the mask areas of $\tilde{\phi}_{d_{i+1}}$ using the region normalization algorithm~\cite{yu2020region}:
\begin{equation}
 \overline{\phi}_{d_{i+1}} = RN(\tilde{\phi}_{d_{i+1}}, m_{i+1}).
\end{equation}
Finally, we pass $\overline{\phi}_{d_{i+1}}$ through a convolutional layer to generate the refined features $\phi_{d_{i+1}}$. Note that we use resized $\hat{M}_{bin}$ as the mask in different refinement blocks.
Multiple such modules are stacked and the final output of our network is up-sampled to the same resolution as the input image. 
This hierarchical generation and refinement process effectively fuses textural and structural information from the deep and shallow layers of the encoder-decoder to generate better face images.

We also introduce a PatchGAN ~\cite{isola2017image} with spectral normalization~\cite{miyato2018spectral} as a discriminator to further improve the visual quality of the inpainted images.
It takes an inpainted image as input and determines whether the image patches of $I_{pred}$ with size of $70\times70$ are real. (For the sake of clarity, the discriminator is not shown in Figure~\ref{figure:framework}).
Additionally, the discriminator also takes the landmarks as input, which could regularize the network to pay more attention to the structurally important regions of human faces. 

\subsection{Objective Functions}
Finally, our inpainting module is trained with a joint loss that consists of a reconstruction loss, an adversarial loss, a perceptual loss, a style loss, and a total variation loss.

\vspace{0.1in}
\noindent \textbf{The reconstruction loss} is defined as follows:
\begin{equation}
    \mathcal{L}_{recons} = \frac{1}{N_{m}} {\left\| I_{pred} - I_{gt} \right\|}_{1}.
\end{equation}
We follow \cite{nazeri2019edgeconnect} to calculate the reconstruction loss only on masked regions and $N_{m}$ is the number of masked pixels. Additionally,  ${\left\| \cdot \right\|}_{1}$ denotes the $\ell_{1}$ norm.

\vspace{0.1in}
\noindent \textbf{The adversarial loss} that we use follows the LSGAN~\cite{mao2017least}, which has demonstrated its ability in stabilizing the training process and improve the visual quality of the images:
\begin{equation}
    \begin{split}
     \mathcal{L}_{adv_{G}} &= \mathbb {E} \left[ (\mathcal{D}(\mathcal{G}_{p}(I_{m_{bin}}, L_{pred}), L_{gt})-1)^{2} \right], \\
     \mathcal{L}_{adv_{D}} &=  \mathbb {E} \left[ (\mathcal{D}(I_{pred}, L_{pred})-1)^{2} \right] +  \mathbb {E} \left[ (\mathcal{D}(I_{gt}, L_{gt})-1)^{2} \right],
    \end{split}
\end{equation}

where $L_{gt}$ denotes the ground-truth landmarks and $\mathcal{D}$ is the discriminator network. As $L_{gt}$ are unavailable for both the CelebA-HQ dataset~\cite{lee2020maskgan} and CelebA dataset~\cite{liu2015faceattributes}, we apply FAN~\cite{bulat2017far} to generate them.

\vspace{0.1in}
\noindent \textbf{The perceptual loss} penalizes the restored results that are not perceptually similar to ground-truth images by measuring the distance between their activation maps:
\begin{equation}
    \mathcal{L}_{perc} = \sum_{p} \frac{{\left\| \phi_{p}(I_{pred}) - \phi_{p}(I_{gt}) \right\|}_{1}}{N_{p} \times H_{p} \times W_{p} },
\end{equation}
where $\phi_{p}(\cdot)$ denotes the $N_{p}$ features maps with size of $H_{p} \times W_{p}$ of the $p_{th}$ layer from the pre-trained network. In this work, $relu1\_1$, $relu2\_1$, $relu3\_1$, $relu4\_1 $ and $relu5\_1$ of the pretrained VGG-19 network are used to calculate the perceptual loss.

\vspace{0.1in}
\noindent \textbf{The style loss} similarly computes the style distance between two images:
\begin{small}
\begin{equation}
    \mathcal{L}_{style} = \sum_{p} \frac{1}{N_{p} \times N_{p}} {\left\| \frac{ G_{p}(I_{pred} \odot \hat{M}_{bin} ) - G_{p}(I_{gt} \odot \hat{M}_{bin} )}{ N_{p} \times H_{p} \times W_{p}} \right\|},
\end{equation}
\end{small}
where $G_{p}(x) = \phi_{p}(x)^\mathrm{T} \phi_{p}(x)$ denotes the Gram Matrix corresponding to $\phi_{p}(x)$. It is validated by \cite{sajjadi2017enhancenet} to be effective in combating the checkerboard effects.

\vspace{0.1in}
\noindent \textbf{The total variation loss} is used to make the restored results visually smoother:
\begin{equation}
    \mathcal{L}_{tv} = \frac{1}{N_{I_{pred}}} {\left\| \nabla I_{pred} \right\|}_{1},
\end{equation}
where $N_{I_{pred}}$ is the number of pixels of image $I_{pred}$, and $\nabla$ denotes the first order derivative, including the horizontal $\nabla_{h}$ and vertical $\nabla_{v}$ directions.

The overall loss is a weighted combination of the above:
\begin{equation}
    \begin{split}
    \mathcal{L}_{inpaiting} =&\lambda_{recons} \mathcal{L}_{recons} + \lambda_{adv} \mathcal{L}_{adv_{G}} +  
    \\&\lambda_{perc} \mathcal{L}_{perc} + \lambda_{style} \mathcal{L}_{style} + \lambda_{tv} \mathcal{L}_{tv}.
    \end{split}
\end{equation}
In this work, we empirically use $\lambda_{recons} = 1$, $\lambda_{adv} = 0.01$, $\lambda_{perc} = 0.1$, $\lambda_{style} = 250$, and $\lambda_{tv} = 0.1$ when training the inpainting module on CelebA-HQ dataset and adjust $\lambda_{recons}$ to 5 on CelebA dataset.

\section{Experiments}
We first compare our method with the state-of-the-art blind inpainting method VCNet~\cite{wang2020vcnet} to evaluate the performance of our complete method. Then we use the ground-truth mask to independently evaluate the TDRefine Inpainting Module under non-blind setting and compare it with state-of-the-art methods.
Finally, we perform ablation studies to validate the contribution of the frequency modality and the Patch Similarity Block in our Transformer-based Mask Detection Module, and the effectiveness of TDRB in our TDRefine Inpainting Module.

\subsection{Experiment Setup}

\noindent \textbf{Datasets.} 
We evaluate our method on the CelebA-HQ dataset~\cite{lee2020maskgan} and CelebA dataset~\cite{liu2015faceattributes}. The mask shapes that we use for training include both randomly generated block masks and free-form strokes adopted from~\cite{yu2019free}. 
When testing, we use the irregular mask dataset~\cite{liu2018image} which has been grouped into six intervals according to the mask area, i.e., 0-10\%, 10-20\%, 20-30\%, 30-40\%, 40-50\%, and 50-60\%, and each interval has 2,000 masks. 
The filling contents in our masks for both training and testing are constant values and ground-truth images from the Places2 dataset~\cite{zhou2017places}. 
All the masks and images for training and testing are resized to 256 $\times$ 256, and the block masks that we generate have the size of 128 $\times$ 128. 

\vspace{0.1in}
\noindent \textbf{Evaluation Metrics.} We apply peak signal-to-noise ratio (PSNR) and structural similarity index metrics (SSIM) as our evaluation metrics, which are common used in painting results evaluation \cite{nazeri2019edgeconnect, li2020recurrent, yang2019lafin,wang2020vcnet}. 

\begin{figure*}[t]
\centering
\includegraphics[width=0.7\linewidth]{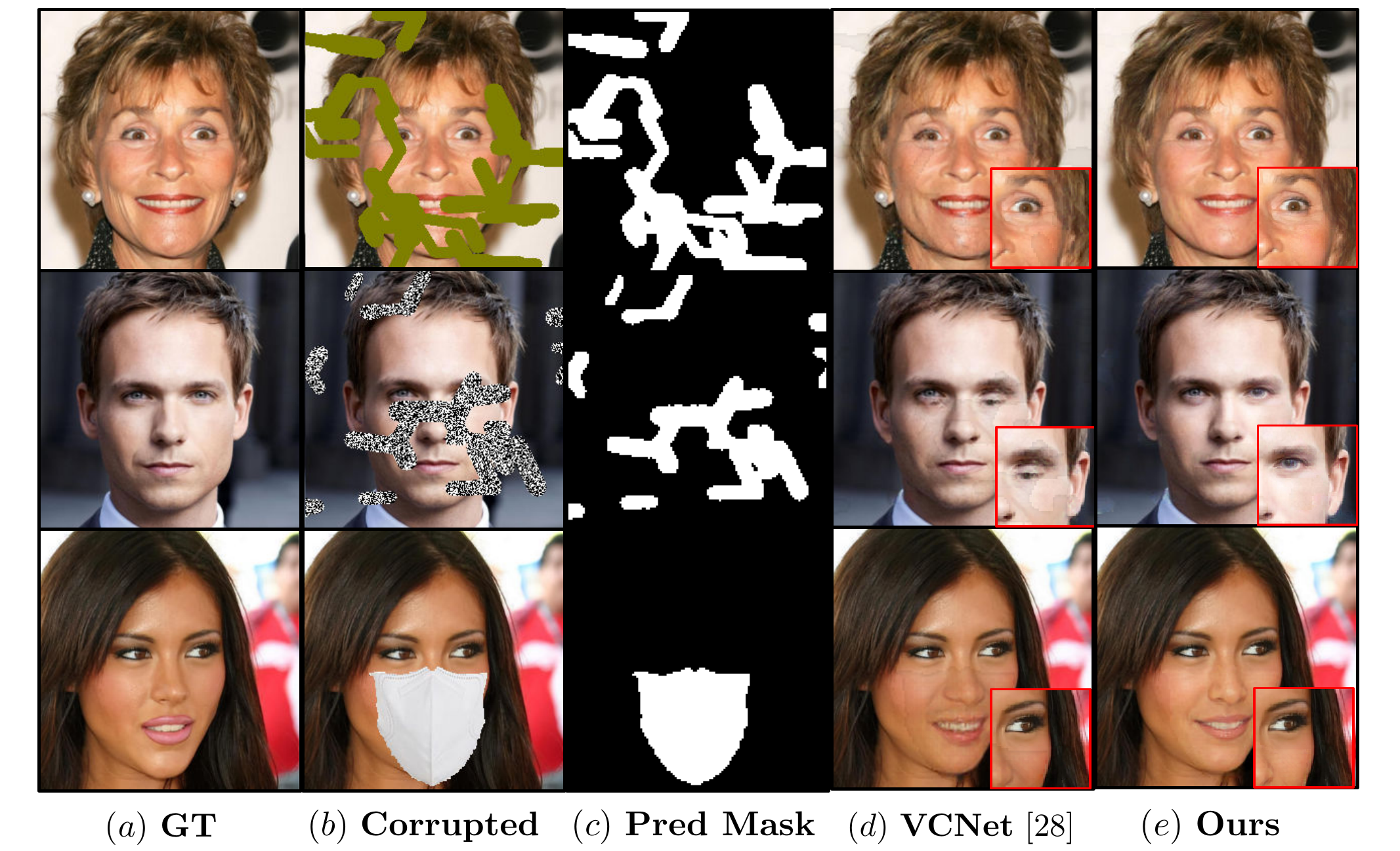} 
\caption{Qualitative comparison with blind inpaining method VCNet \cite{wang2020vcnet} on the CelebA-HQ dataset. From left to right, we demonstrate (a) the ground-truth image, (b) the corrupted image, the inpainting results from (c) the predicted mask of our method, (d) the inpainting results of VCNet and (d) the inpainting results of our method, respectively.}
\label{figure:blind}
\end{figure*}

\begin{figure}[t]
  \centering
  \includegraphics[width=\linewidth]{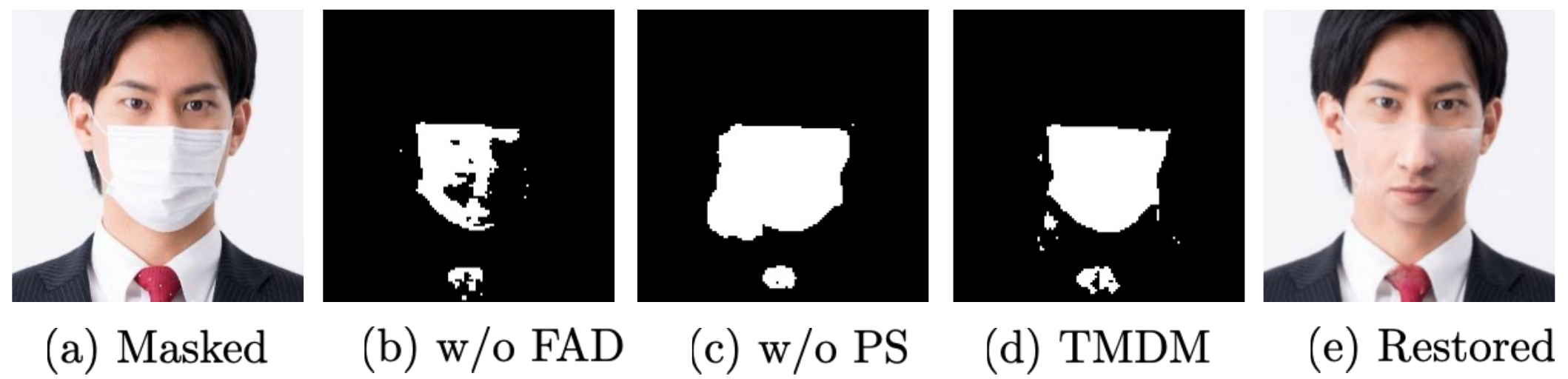}
  \caption{Visualization of the examples of mask detection with/without the PS block and FAD. From left to right, we show the corrupted images, mask detection module without FAD, mask detection module without the PS block, the detected masks using our TMDM, and the restored face image using our predicted mask, respectively.}
  \label{fig:ps}
\end{figure}

\vspace{0.1in}
\noindent \textbf{Implementation Details.} 
We conduct experiments to evaluate the performance of our method under both blind inpainting setting and non-blind inpainting setting. For blind face inpainting, we use a two-stage training strategy. First, we train the mask detection module $\mathcal{G}_{M}$ independently. Then we train the mask detection module and inpainting module $\mathcal{G}_{P}$ jointly in an end-to-end manner. The batch sizes are 16 and 8 for the first and second stages, respectively. For non-blind face inpainting, we independently train the inpainting module $\mathcal{G}_{P}$ using ground-truth masks.
The learning rates of the generator and discriminator in $\mathcal{G}_{P}$ are set to $10^{-4}$ and $10^{-5}$, respectively. The learning rate of $\mathcal{G}_{M}$ is set to $10^{-4}$. All the parameters are optimized by the Adam optimizer with $\beta_{1} = 0$ and $\beta_{2}=0.9$.

\subsection{Blind Inpainting Evaluation}

\subsubsection{Mask Prediction Results} We first evaluate the mask detection performance using MAE loss (lower is better) and IoU (higher is better) on the irregular mask dataset, and report the quantitative results in Table~\ref{table:mask}. As the table shows, our transformer-based mask detection module can accurately predict the masks of different sizes. Further analysis on the mask detection module will be provided in Section~\ref{sec:ablation}.

\begin{table}[t]
\caption{Mask detection results for different mask areas on the irregular mask dataset.}
\label{table:mask}
\centering
\setlength{\tabcolsep}{1.6mm}{
\begin{tabular}{lccccc}
\toprule
     \textbf{Metric} && 10-20\% & 20-30\% & 30-40\% & 40-50\%  \\
\cmidrule{1-1} \cmidrule{3-6}
     \textbf{MAE} && 1.93 & 2.01 & 2.13 & 2.24  \\
     \textbf{IoU} && 91.04\% & 92.37\% & 94.17\% & 96.33\% \\
\bottomrule

\end{tabular}}
\end{table}

\begin{table}[t]
\caption{Quantitative results on the CelebA-HQ dataset for: VCNet \cite{wang2020vcnet}, and our method. $^{\star}$ denotes inpainting with predicted masks from $\mathcal{G}_{M}$. For both PSNR and SSIM, higher is better.}

\label{table:blind}
\centering

\setlength{\tabcolsep}{2mm}{
		\begin{tabular}{lccccc}
			\toprule
			\textbf{Metric} && \textbf{Mask} && VCNet \cite{wang2020vcnet} & Ours$^{\star}$ \\
			\cmidrule{1-1} \cmidrule{3-3} \cmidrule{5-6}
			\multirow{3}*{PSNR} && 10-20\% && 30.82  & \textbf{31.61}\\
			~ && 40-50\% && 24.11  & \textbf{24.39}\\
			~ && Center &&  25.47 & \textbf{26.01}\\
			\cmidrule{1-1} \cmidrule{3-3} \cmidrule{5-6}
			\multirow{3}*{SSIM} && 10-20\% && 0.969 & \textbf{0.974}\\
			~ && 40-50\% && 0.867 & \textbf{0.877}\\
			~ && Center && 0.883 & \textbf{0.908}\\
			\bottomrule
	\end{tabular}}
\end{table}

\subsubsection{Inpainting Results Comparison}  

\vspace{0.2in}
\noindent \par \textbf{Quantitative Comparisons.} Previous blind inpainting methods \cite{cai2017blind,liu2019deep} have not published their code, therefore, we compare our method with the most recent state-of-the-art method, VCNet~\cite{wang2020vcnet}, and report the results in Table~\ref{table:blind}. The results clearly demonstrate that our method outperforms VCNet for various types of masks applied to the face images, i.e., about 2.6\%, 1.2\%, and 2.1\% performance gain in PSNR score on the 10-20\%, 40-50\%, and Center masks, respectively. 

\vspace{0.1in}
\noindent \textbf{Qualitative Comparisons.}
We further compare the inpainting results of VCNet \cite{wang2020vcnet} and our method qualitatively and show some examples in Figure~\ref{figure:blind}. We can see that the restored contents by VCNet and our method are basically visually reasonable, but our proposed method can produce the most visually natural and pleasing details which are consistent with the surrounding context. 
Except from the high-quality mask predictions, this is mainly because 1) the multiple refinement fusion blocks fuse the information at different levels more effectively and 2) the generator also uses proper structural priors as guidance.

\subsection{Non-blind Inpainting Evalution}
\noindent \textbf{Quantitative Comparisons.} 
To specifically demonstrate the effectiveness of our inpainting module,
we compare our method with state-of-the-art non-blind inpainting methods: EC \cite{nazeri2019edgeconnect}, RFR \cite{li2020recurrent}, and Lafin \cite{yang2019lafin} on the CelebA-HQ dataset and report the results in Table~\ref{table:cel-hq}.
It can be seen that our method achieves 31.75, 28.75, 26.40, 24.45, 21.75, and 26.16 in PSNR on the 10-20\%, 20-30\%, 30-40\%, 40-50\%, 50\%+, and Center masks, respectively, which outperforms the current state-of-the-art methods.
In addition, further quantitative comparisons with CE~\cite{pathak2016context}, EC~\cite{nazeri2019edgeconnect}, andLafin~\cite{yang2019lafin} on center masks on the CelebA dataset are shown in Table~\ref{table:cel}. 
We also achieve the best results.

\begin{table}[t]
\caption{Quantitative results on the CelebA-HQ dataset for: EC \cite{nazeri2019edgeconnect}, RFR \cite{li2020recurrent}, Lafin \cite{yang2019lafin} and our method. For both PSNR and SSIM, higher is better.}
	
	\label{table:cel-hq}
	
	\centering
	\resizebox{.85\columnwidth}{!}{
		\begin{tabular}{lccccccc}
			\toprule
			\textbf{Metric} && \textbf{Mask} && EC \cite{nazeri2019edgeconnect} & RFR \cite{li2020recurrent} & Lafin \cite{yang2019lafin}  & Ours\\
			\cmidrule{1-1} \cmidrule{3-3} \cmidrule{5-8}
			\multirow{6}*{\rotatebox{90}{PSNR}} && 10-20\% && 30.73 & 30.92 & 31.48 & \textbf{31.75}\\
			~ && 20-30\%  && 27.56 & 28.02 & 28.31 & \textbf{28.57}\\ 
			~ && 30-40\%  && 25.34 & 25.79 & 26.14 & \textbf{26.40}\\
			~ && 40-50\% && 23.44 & 23.96 & 24.22 & \textbf{24.45}\\
			~ && 50\%+ && 20.71 & 21.33 & 21.61 & \textbf{21.75}\\
			~ && Center && 24.82 & 25.47 & 25.92 & \textbf{26.16}\\
			\cmidrule{1-1} \cmidrule{3-3} \cmidrule{5-8}
			\multirow{6}*{\rotatebox{90}{SSIM}} && 10-20\% && 0.971 & 0.972 & 0.975 & \textbf{0.978}\\
			~ && 20-30\%  && 0.942 & 0.948 & 0.951 & \textbf{0.959}\\ 
			~ && 30-40\% && 0.907 & 0.915 & 0.922 & \textbf{0.932}\\
			~ && 40-50\% && 0.859 & 0.870 & 0.883 & \textbf{0.885}\\
			~ && 50\%+ && 0.754 & 0.773  & 0.805 &  \textbf{0.811}\\
			~ && Center && 0.874 & 0.883 & 0.905 & \textbf{0.912}\\
			\bottomrule
	\end{tabular}}

\end{table}

\begin{table}[t]
\caption{Quantitative results on the CelebA dataset in terms of PSNR and SSIM on center masks for: CE~\cite{pathak2016context}, EC~\cite{nazeri2019edgeconnect}, Lafin~\cite{yang2019lafin}, and Ours.}
\label{table:cel}
	\centering
	\footnotesize
	\resizebox{.75\columnwidth}{!}{
	
	\begin{tabular}{lcccccc}
		\toprule
		\textbf{Metric}  && CE~\cite{pathak2016context} & EC~\cite{nazeri2019edgeconnect} & Lafin~\cite{yang2019lafin}  & Ours\\
		\cmidrule{1-1} \cmidrule{3-6}
		PSNR &&  25.46 & 25.83 & 26.25 & \textbf{26.28} \\
		SSIM && 0.909 & 0.899 & 0.912 & \textbf{0.917}\\
		\bottomrule
	\end{tabular}}

\end{table}

\begin{table}[t]
\caption{Quantitative results of the variants of our method, we remove FAD or PS Block in $\mathcal{G}_{M}$ for mask prediction. The experiments are conducted on the irregular mask dataset.}
\label{table:abl1}
	\centering
	\resizebox{.95\columnwidth}{!}
	{\begin{tabular}{lccccccc} 
		\toprule
		\textbf{Metric} && \textbf{Mask} && w/o DA & w/o FAD & w/o PS & Ours   \\
		\cmidrule{1-1} \cmidrule{3-3} \cmidrule{5-8}
		\multirow{2}*{MAE} && 10-20\% && 2.58 & 2.13 & 2.21 &  1.93 \\
		~ && 40-50\% && 2.72 & 2.41 & 2.47 &  2.24\\ 
		\cmidrule{1-1} \cmidrule{3-3} \cmidrule{5-8}
		\multirow{2}*{IoU} && 10-20\% && 85.39\% & 89.16\% & 89.55\% &  91.04\%  \\
		~ && 40-50\% && 92.18\%  & 95.06\% 91.& 95.32\% & 96.33\% \\ 
		\bottomrule
	\end{tabular}}
\end{table}

\begin{figure*}[t]
\centering
\includegraphics[width=0.75\linewidth]{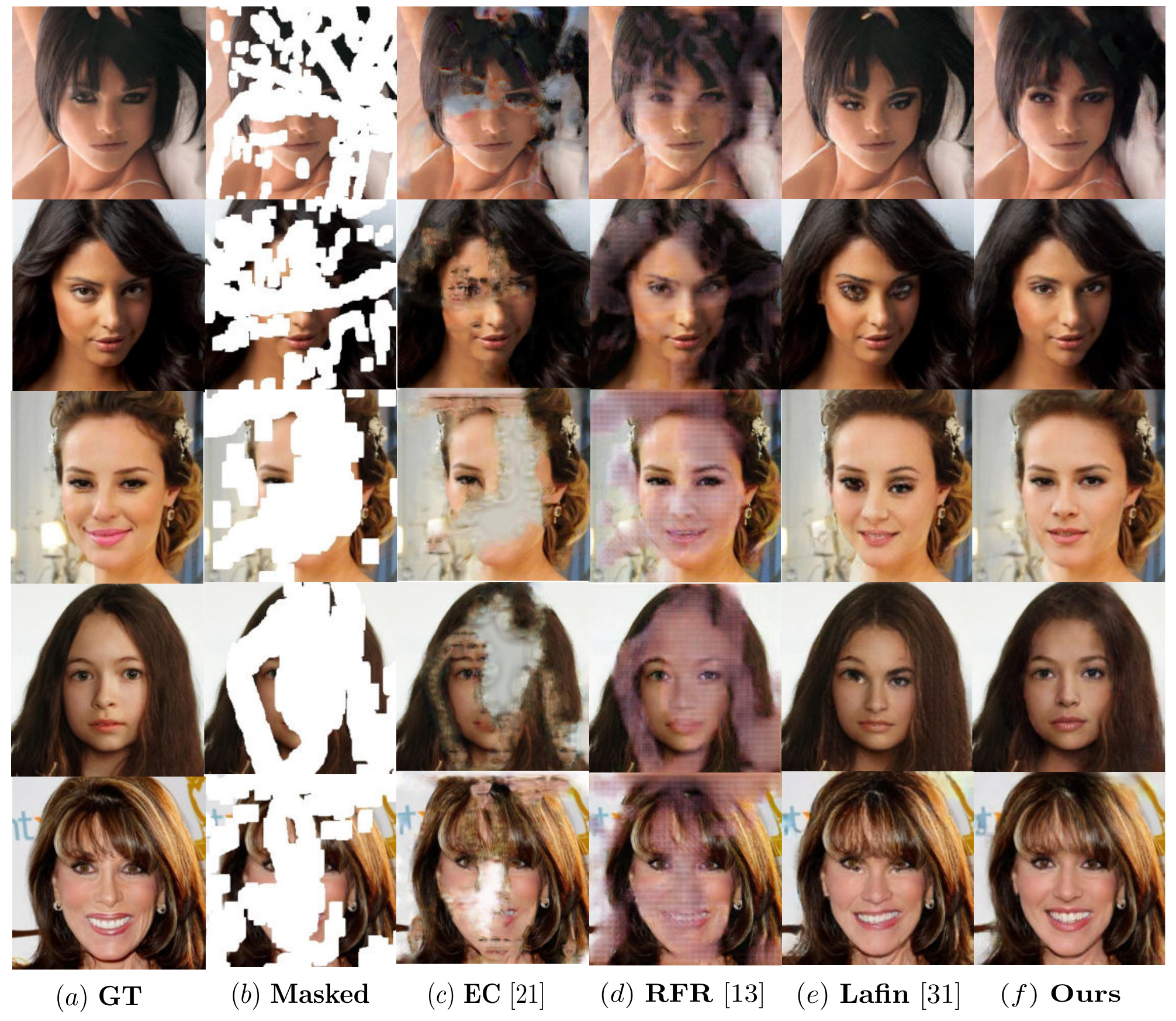} 
\caption{Qualitative comparison with other state-of-the-art face inpainting methods on the CelebA-HQ dataset. From left to right, we demonstrate (a) the ground-truth image, (b) the masked image, and the inpainting results from (c) EC, (d) RFR, (e) Lafin, and (f) our method with the predicted mask, respectively.}
\label{figure:non-blind}
\end{figure*}

\begin{figure}[t]
\centering
	\includegraphics[width=0.8\columnwidth]{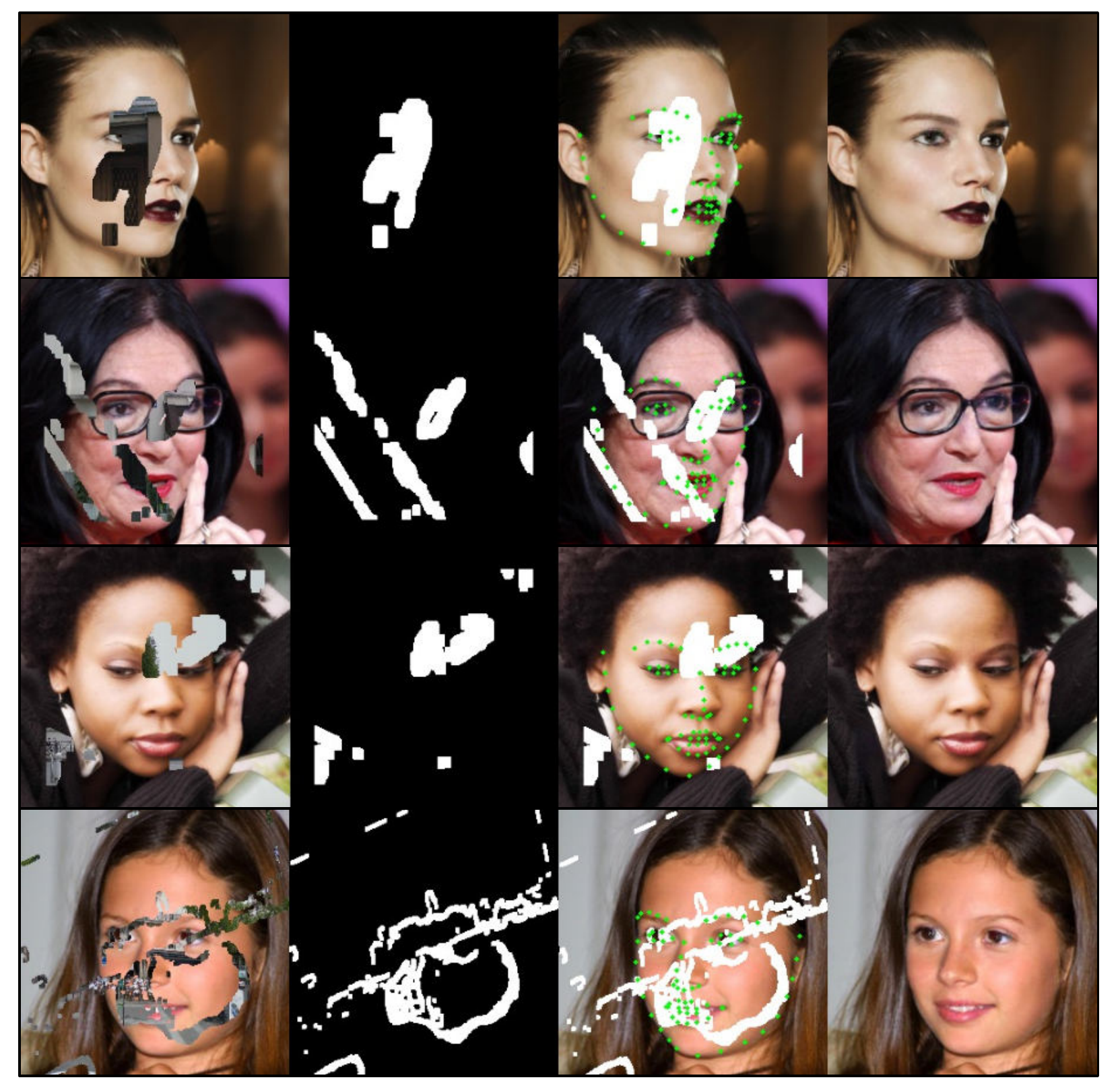}
	\caption{Face inpainting results by our proposed blind face inpainting method on CelebA-HQ dataset. From left to right, we show the corrupted face images, the predicted binary masks, the predicted landmarks on corrupted images, and the restored results, respectively. } 
\label{fig:landmark}
\end{figure}

\vspace{0.1in}
\noindent \textbf{Qualitative Comparisons.} 
We present qualitative results of our method and state-of-the-art methods in Figure~\ref{figure:non-blind}, 
which shows the images inpainted by EC~\cite{nazeri2019edgeconnect}, RFR~\cite{li2020recurrent}, Lafin~\cite{yang2019lafin}, and ours (using ground-truth masks) on the CelebA-HQ dataset. 
Note that for EC, the pre-trained models on CelebA-HQ dataset are not provided by the authors, so we use the code that they provide to train on the CelebA-HQ dataset by ourselves. 
It can be seen that EC \cite{nazeri2019edgeconnect} and RFR~\cite{li2020recurrent} generate blurred results when the faces have rich expressions and postures because they do not use suitable structural guidance to facilitate face restoration. 
Just as \cite{yang2019lafin} suggests, the redundancy of edge information may even degrade the performance.
Lafin \cite{yang2019lafin} mitigate this problem to some extent but struggle to preserve the properties of facial attributes, e.g., eyes and mouths.
Comparatively, we tackle the problem by the refinement fusion blocks to hierarchically restore features, and our method generates the most natural and visually pleasant contents.

\subsection{Ablation Study}\label{sec:ablation}

\noindent \textbf{Mask detection results w/o FAD and PS block.}
In addition, we show the mask detection results on face images wearing real masks without FAD and PS block in Figure~\ref{fig:ps}. The comparison results demonstrate that using PS block and FAD enables our mask detection module to not only accurately detect the visually abnormal regions on face images, but also develop sharp capability in edge perception. 

\vspace{0.1in}
\noindent \textbf{Effects of different components in TMDM.} The Frequency Anomaly Detector (FAD) is utilized to detect the visually abnormal signals of face images in the frequency domain, while the Dual Attention (DA) combines the self-attention map and the frequency modality attention map to locate the masked regions. To evaluate the effects of FAD and DA, we separately remove them from TMDM and show the performance degradation in Table~\ref{table:abl1}. In addition, we also remove the Patch Similarity Block to validate its contribution in mask detection. The quantitative comparison results validates that different components can effectively improve the performance of our mask detection module by a large margin, \eg, the use of DA, FAD and PS increases the IoU values by about 6.6\%, 2.1\% and 1.7\% on the 10-20\% masks.

\begin{table}[t]
\caption{Quantitative results on the CelebA-HQ dataset in terms of PSNR and SSIM on center masks. DC denotes only using Deconv + 1$\times$1 Conv blocks for upsampling, and DCC denotes using Deconv + 1$\times$1 Conv blocks for upsampling and Concatenation for naive fusion.}
\label{table:abl2}
	\centering
	\footnotesize
	\resizebox{.55\columnwidth}{!}{
		\begin{tabular}{lcccc}
			\toprule
			\textbf{Metric}  && DC & DCC  & Ours\\
			\cmidrule{1-1} \cmidrule{3-5}
			PSNR &&  24.85 & 25.32 & \textbf{26.16}\\
			
			SSIM && 0.881 & 0.893 & \textbf{0.912}\\
			\bottomrule
	\end{tabular}}
\end{table}

\vspace{0.1in}
\noindent \textbf{Effects of the TDRB.} We demonstrate the contributions of top-down refinement fusion blocks (TDRB) by  1) replacing them with (Deconv+1$\times$1 Conv) blocks for up-sampling 2) using (Deconv+1$\times$1 Conv) blocks for up-sampling and concatenation for naive fusion. 
The quantitative results in Table~\ref{table:abl2} show that both variants achieve lower performances (i.e., 5.0\% and 3.2\% performance degradation in terms of PSNR value, respectively),
this validates that our refinement fusion block can  improve the quality of output images by effectively merging the structural information from deep layers with textural information from shallow layers.

\vspace{0.3in}

\noindent \textbf{Landmark prediction results on masked images.}
For our proposed inpainting method, the landmark prediction module is just a tool to provide geometry information to the generator, instead of our main contribution. Therefore, we adopted the method of Lafin~\cite{yang2019lafin}. We demonstrate the performance of the landmark prediction module below (Figure~\ref{fig:landmark}), and the results validate it could detect landmarks accurately even on the damaged images. The restored results by \system are also shown in the  rightmost column.

\begin{figure}[t]
  \centering
  \includegraphics[width=0.9\linewidth]{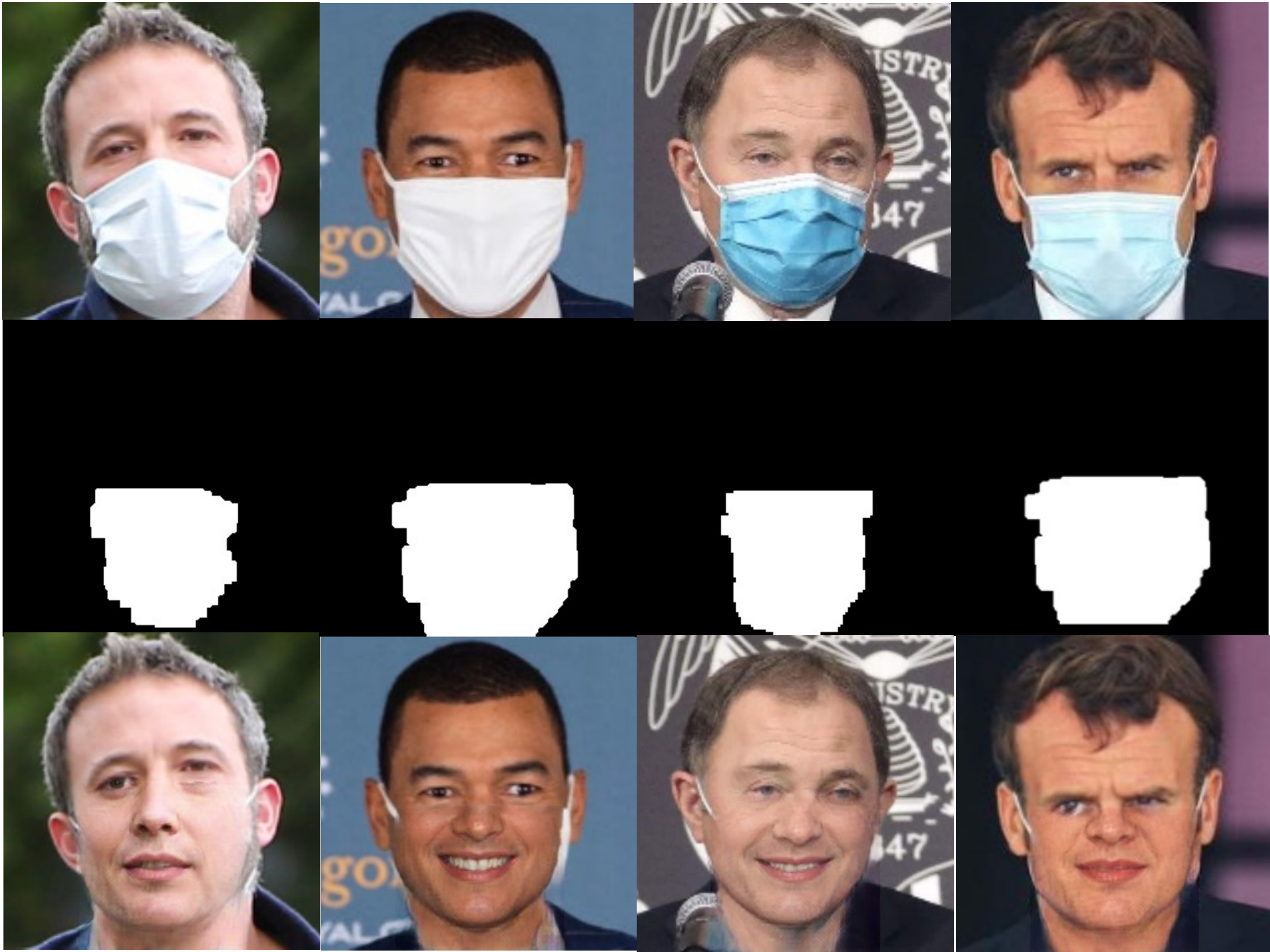}
  \caption{Visualisation results of the proposed blind face inpainting method. From top to bottom, we demonstrate the masked face images, the predicted masks, and the restored results (with the predicted masks), respectively.}
  \label{fig:real}
\end{figure}

\vspace{0.1in}
\noindent \textbf{Evalutaion on the In-the-wild data.}
In order to further validate the effectiveness of \system on real-world masked images, we evaluate our method on a public masked face image dataset MFR2~\cite{anwar2020masked}, and demonstrate the predicted masks and inpainted results in Figure~\ref{fig:real}. Note that both ground-truth masks and ground-truth uncorrupted face images are not available. We can observe that although the patterns of these masks are not included in our training data, \system can still locate the mask regions, and the restored contents are visually reasonable.

\begin{figure}[t]
  \centering
  \includegraphics[width=0.9\linewidth]{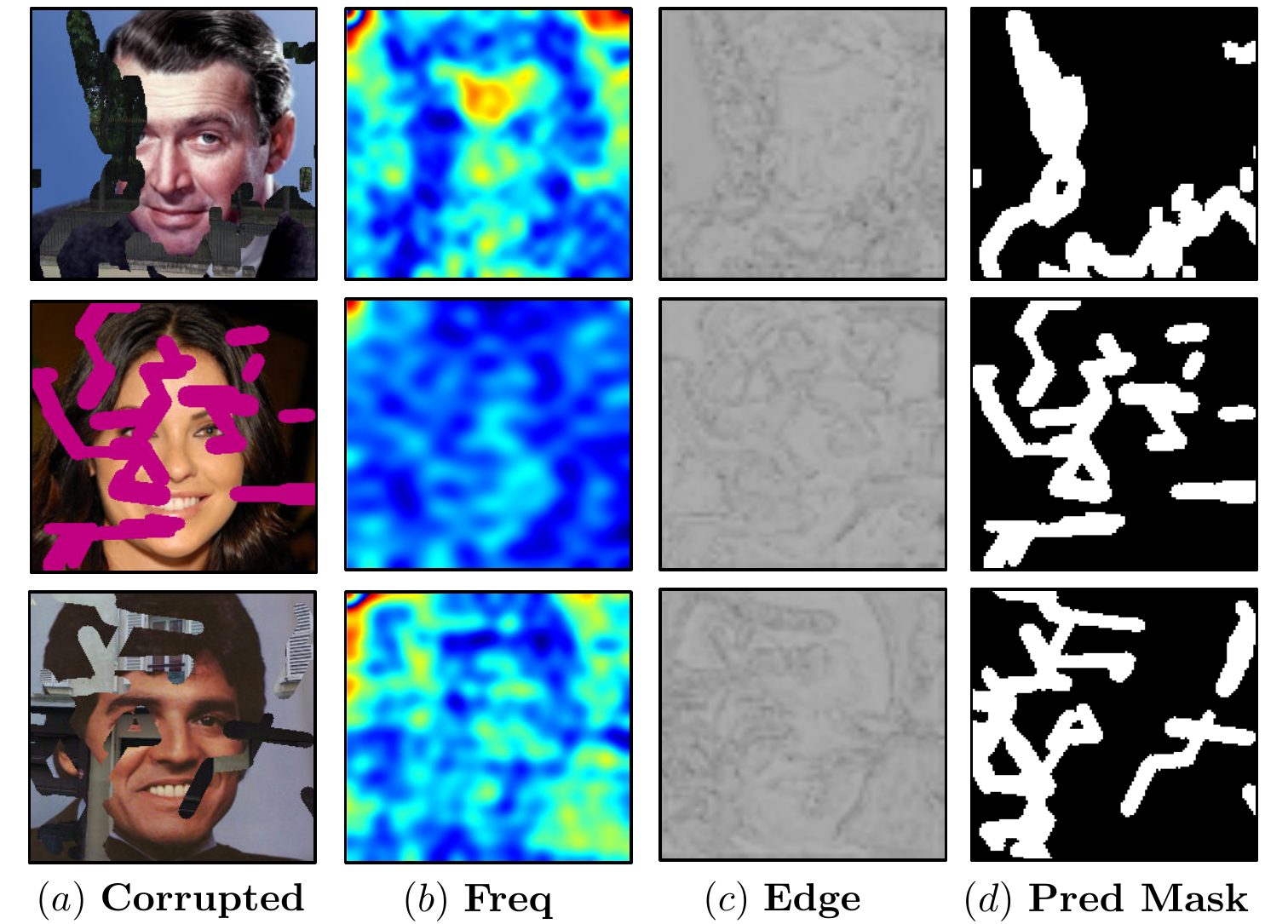}
  \caption{Visualization of (a) the corrupted image, (b) the frequency attention $A_{freq}$, (c) the edge map $E$, and (d) the predicted mask, respectively.}
  \label{fig:vis}
\end{figure}

\vspace{0.1in}
\noindent \textbf{Visualization of the frequency attention and edge map.}
In addition, we visualize the frequency attention $A_{freq}$ and the edge map $E$ produced by the PS Block in Figure \ref{fig:vis}. 
The visualization results demonstrate that the frequency information can roughly locate the visually abnormal regions,
while the edge information can provide additional assistance to develop sharp capability in edge perception.

\vspace{0.1in}
\noindent \textbf{Evaluation of the identity retention capability.} In order to assess the identity retention capability of different face inpainting methods, we propose to leverage ICS to compute the cosine similarity of the features extracted from the restored images and the ground-truth images. ICS is defined as:
\begin{equation}
    ICS = Sim(f(I_{gt}), f(I_{pred})).
\end{equation}
where $Sim$ denotes the cosine similarity function, and the feature extraction network that we use is Inception-V3~\cite{szegedy2015rethinking}. We report the comparison results with state-of-the-art face inpainting methods in Table~\ref{table:ics}, which demonstrate that the restored results by \system could effectively preserve identity information.

\begin{table}[t]
\caption{ICS values of different face inpainting methods on CelebA-HQ dataset. Note that "Ours" denotes restoring the face images with ground-truth masks, while "Ours$^{*}$" denotes using our predicted masks.}
\label{table:ics}
	\centering
	\resizebox{.9\columnwidth}{!}{
		\begin{tabular}{lccccccc}
			\toprule
			\textbf{Metric} && \textbf{Mask} && EC \cite{nazeri2019edgeconnect} & Lafin \cite{yang2019lafin}  & Ours & Ours$^{*}$\\
			\cmidrule{1-1} \cmidrule{3-3} \cmidrule{5-8}
			\multirow{3}*{\rotatebox{90}{ICS}} && 20-30\% && 0.88 & 0.93 & \textbf{0.97} & 0.935\\
			~ && 40-50\% && 0.83 & 0.86 & \textbf{0.89} & 0.86 \\
			~ && Center && 0.87 & 0.89 & \textbf{0.94} & 0.93 \\
			\bottomrule
	\end{tabular}}
\end{table}

\begin{figure}[t]
\centering
		\subfigure{
			\includegraphics[width=0.9\columnwidth]{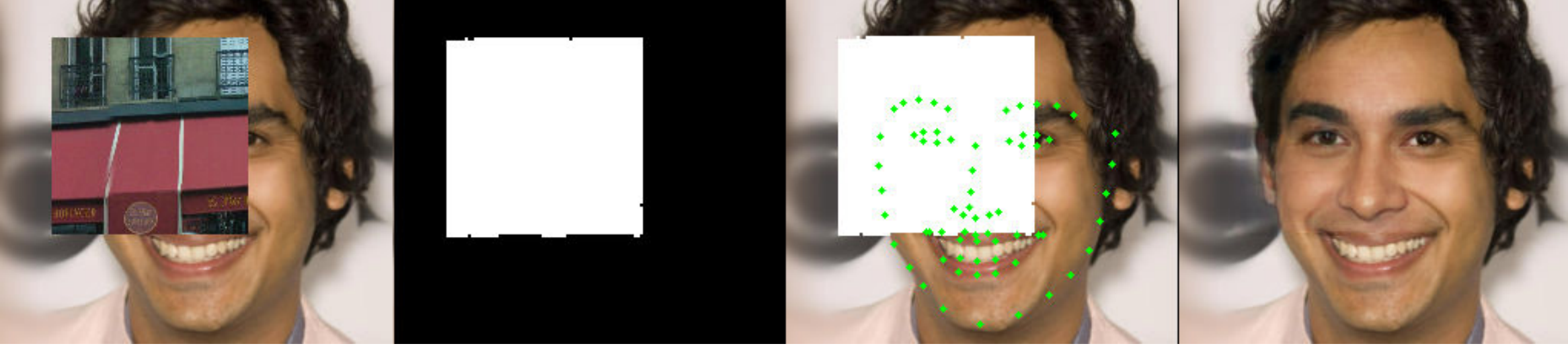}
		}\vspace{-2mm}
		\subfigure{
			\includegraphics[width=0.9\columnwidth]{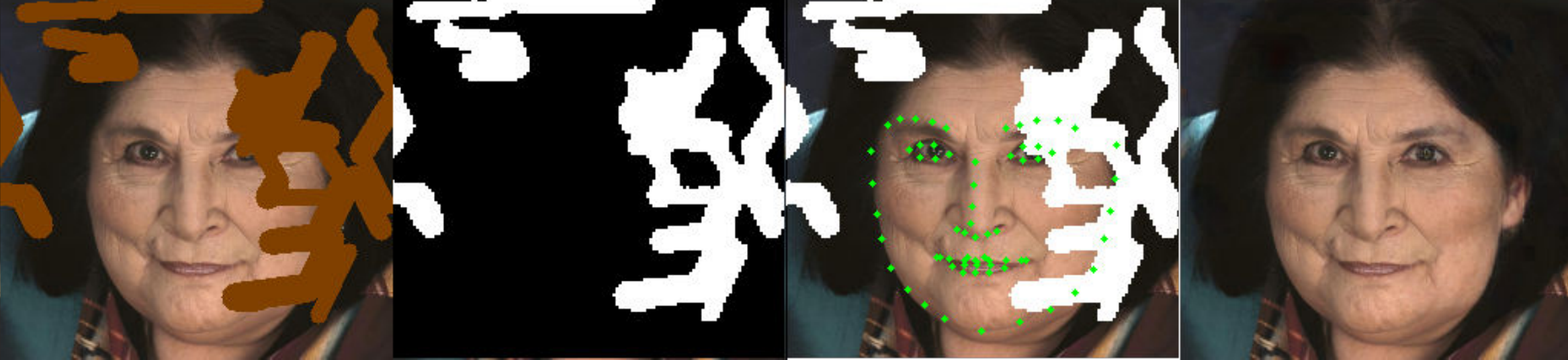}     
		}\vspace{-2mm}
		\subfigure{
			\includegraphics[width=0.9\columnwidth]{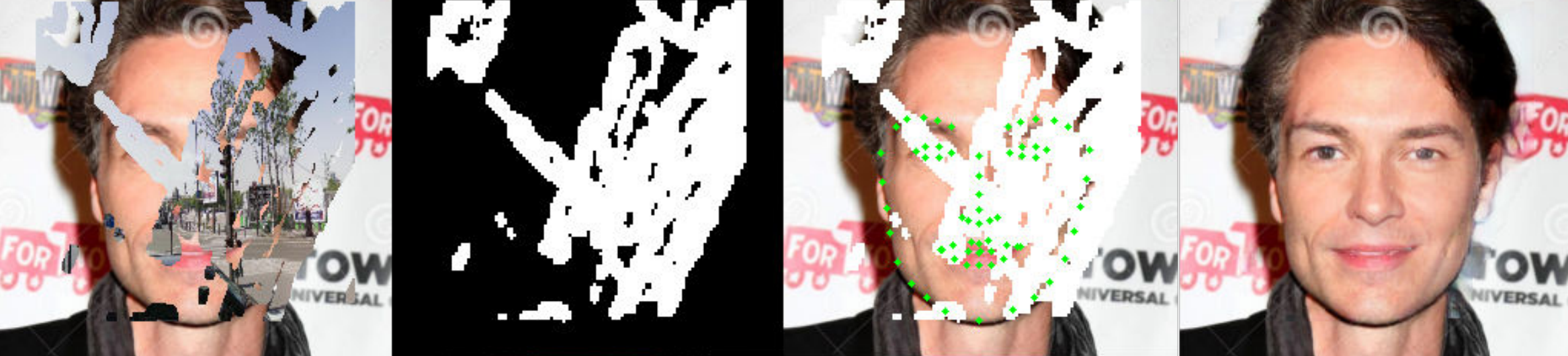}     
		}\vspace{-2mm}
		\subfigure{
			\includegraphics[width=0.9\columnwidth]{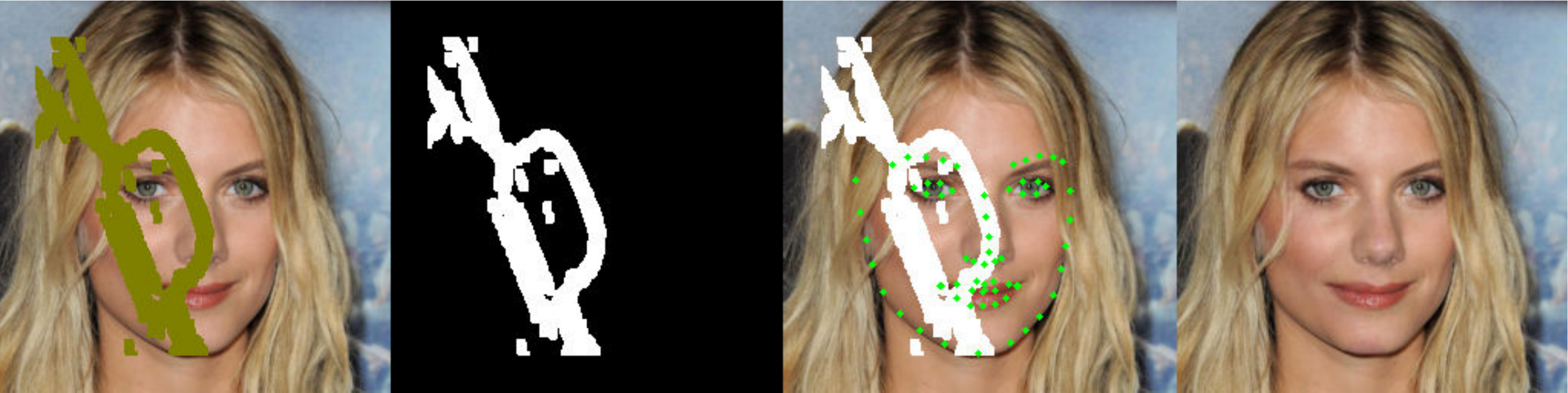}     
		}\vspace{-2mm}
		\subfigure{
			\includegraphics[width=0.9\columnwidth]{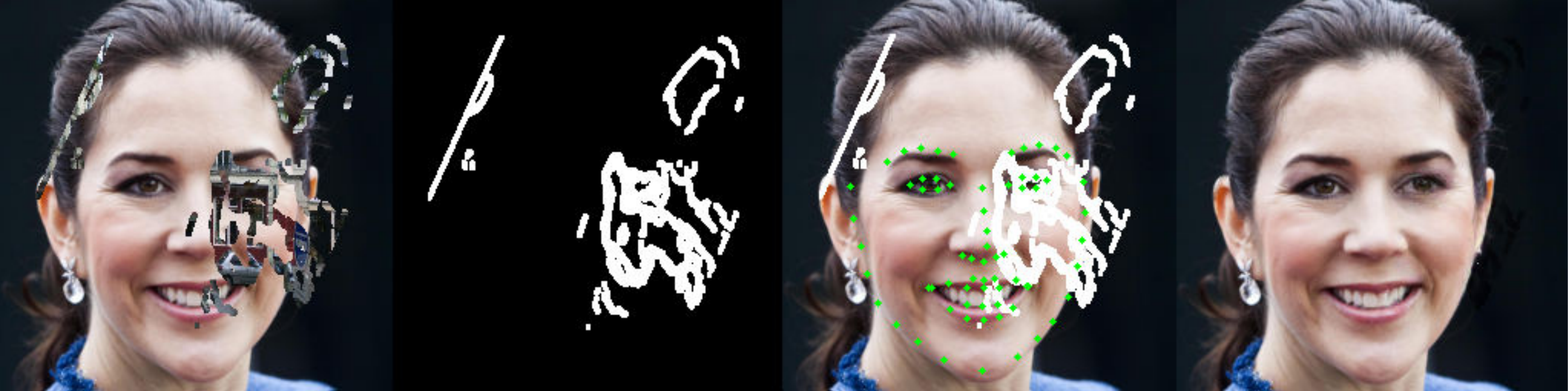}     
		}
	\caption{Face inpainting results by our proposed blind face inpainting method on CelebA-HQ dataset. From left to right, we show the corrupted face images, the predicted binary masks, the predicted landmarks on corrupted images, and the restored results, respectively. } 
\label{fig:supp1}
\end{figure}

\begin{figure}[t]
		\subfigure{
			\includegraphics[width=0.95\columnwidth]{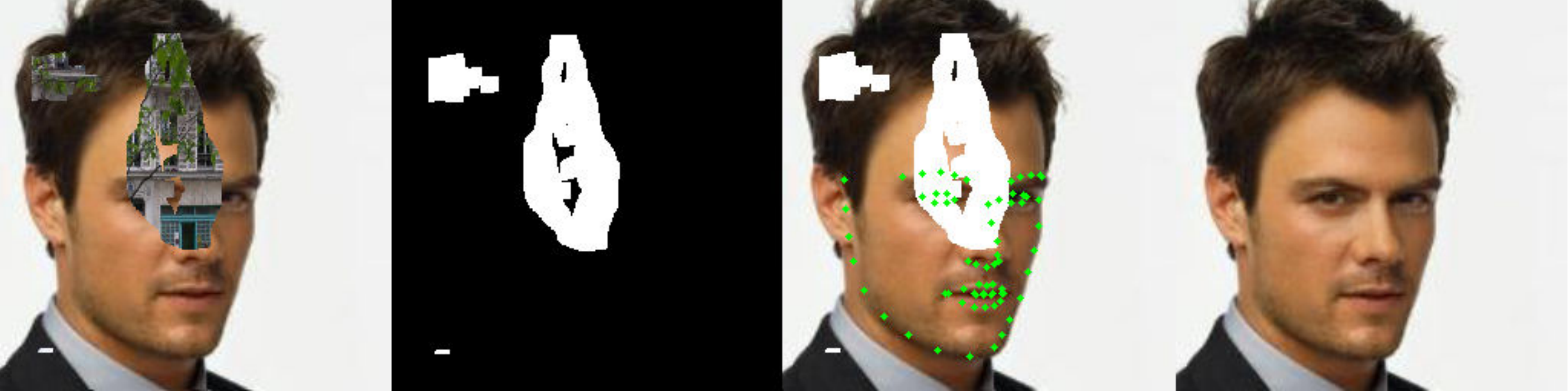}
		}\vspace{-2mm}
		\subfigure{
			\includegraphics[width=0.95\columnwidth]{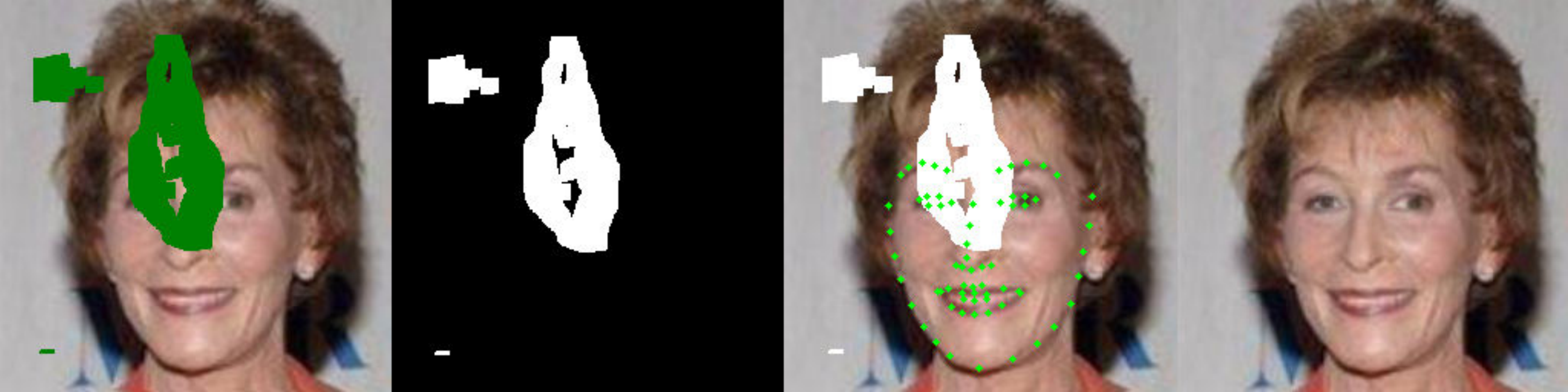}     
		}\vspace{-2mm}
		\subfigure{
			\includegraphics[width=0.95\columnwidth]{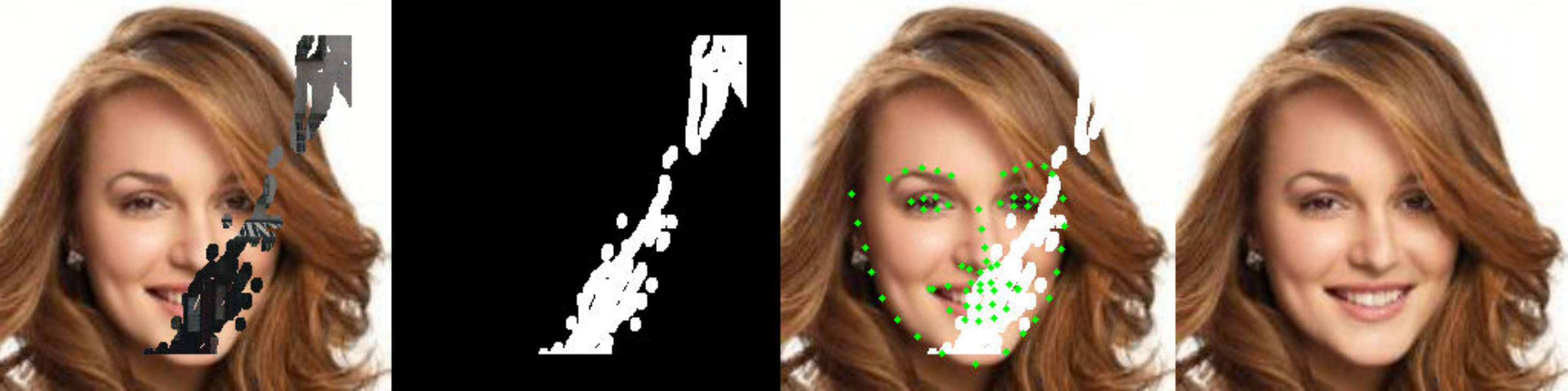}     
		}\vspace{-2mm}
	\caption{Face inpainting results by our proposed blind face inpainting method on CelebA dataset. From left to right, we show the corrupted face images, the predicted binary masks, the predicted landmarks on corrupted images, and the restored results, respectively. } 
\label{fig:supp2}
\end{figure}

\vspace{0.1in}
\noindent \textbf{Additional qualitative results.}
We provide additional results produced by our model on the CelebA-HQ and CelebA datasets in Figure~\ref{fig:supp1} and Figure~\ref{fig:supp2}, respectively.

\section{Discussion}
Benefited from the transformer's capability to capture contextual inconsistencies, our mask detection module can identify unusual visual areas such as graffiti and masks that do not appear on most face images. However, one remaining challenge and limitation of blind inpainting methods (including ours) is that it is difficult to recognize small items like beards and jewelry. 
This is in part due to the dataset bias.
Under the blind inpainting setting, original images in the face image datasets are defined as uncorrupted face images. However, some uncorrupted face images can also contain abnormal visual areas like caps and beards and these won't be marked as masks during the training process, thus the resulting model is not able to complete these areas.
Such dataset bias may be addressed in the future by constructing a cleaner dataset for face inpainting.

\section{Conclusion}
In this work, we propose a new method \system for the task of blind face inpainting, which completes visual contents on a corrupted face image without specifying the damaged regions. Our method first accurately detects the corrupted region and then fills these regions with coherent contents. Specifically, the proposed transformer-based mask detection module operates on image patches in a sequence-to-sequence manner, it also incorporates frequency modality information and captures contextual inconsistency among the patches.
Then in the image generation stage, an encoder-decoder generator with a stack of top-down refinement blocks is used to hierarchically restore features within the masked regions. Texture and structure information is properly combined in the bottom-up and top-down paths.
Extensive experimental results on the widely used CelebA-HQ and CelebA datasets have demonstrated our proposed model can outperform state-of-the-art face image inpainting methods with both ground-truth and predicted masks.

\section{Acknowledgment}
The authors would like to thank the anonymous referees for their valuable comments and helpful suggestions. This work was supported in part by NSFC project (\#62032006).

\vspace{1in}

\bibliographystyle{IEEEtran}
\bibliography{main}

\vspace{4in}

\begin{IEEEbiography}[{\includegraphics[width=1.2in,height=1.25in,clip,keepaspectratio]{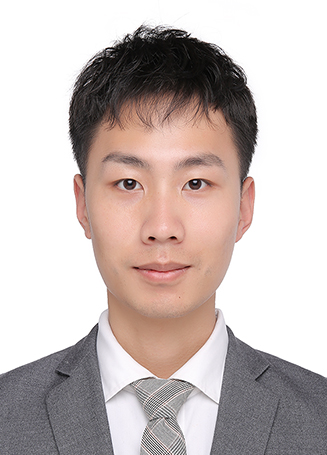}}]{Junke Wang} 
received the B.E. degree from Fudan University, Shanghai, China, in 2021. He is currently pursuing his Ph.D. degree in Computer Science at Fudan University. His research interests include video understanding and media forensics.
\end{IEEEbiography}

\vspace{-3in}

\begin{IEEEbiography}[{\includegraphics[width=1.2in,height=1.25in,clip,keepaspectratio]{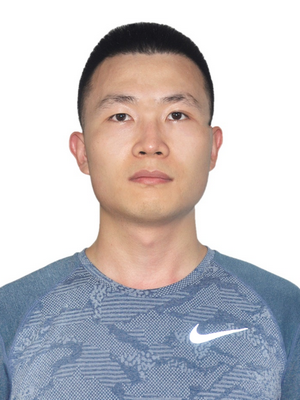}}]{Shaoxiang Chen} is currently a PhD student in the School of Computer Science of Fudan University. Shaoxiang received his B.S. degree from the School of Computer Science of Fudan University. His research is focused on multimedia and deep learning, with respect to video captioning and temporal sentence localization in videos.
\end{IEEEbiography}
\vspace{-3in}

\begin{IEEEbiography}[{\includegraphics[width=1.2in,height=1.25in,clip,keepaspectratio]{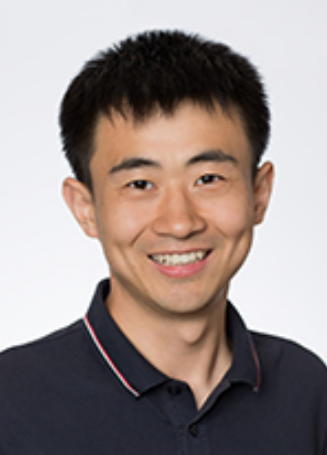}}]{Zuxuan Wu} 
received his Ph.D. in Computer Science from the University of Maryland with Prof. Larry
Davis in 2020. He is currently an Associate Professor in the School of Computer Science at Fudan
University. His research interests are in computer vision and deep learning. His work has been recognized by an AI 2000 Most Influential Scholars Honorable Mention in 2021, a Microsoft Research PhD Fellowship in 2019 and a Snap PhD Fellowship in 2017.
\end{IEEEbiography}
\vspace{-3in}

\begin{IEEEbiography}[{\includegraphics[width=1.2in,height=1.25in,clip,keepaspectratio]{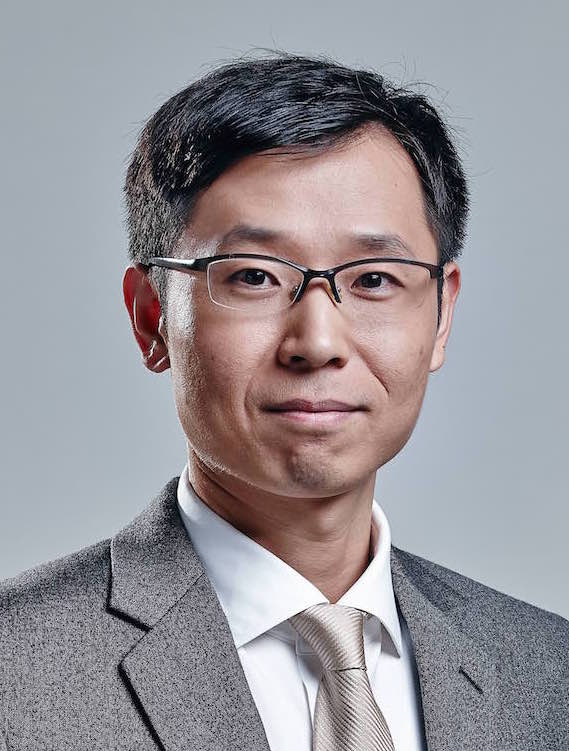}}]{Yu-Gang Jiang} received the Ph.D. degree in Computer Science from City University of Hong Kong in 2009 and worked as a Postdoctoral Research Scientist at Columbia University, New York during 2009-2011. He is currently a Professor and Dean at School of Computer Science, Fudan University, Shanghai, China. His research lies in the areas of multimedia, computer vision and trustworthy AI. His work has led to many awards, including the inaugural ACM China Rising Star Award, the 2015 ACM SIGMM Rising Star Award, and the research award for excellent young scholars from NSF China. 
\end{IEEEbiography}

\end{document}